\def\year{2021}\relax
\documentclass[letterpaper]{article} 
\usepackage{aaai21}  
\usepackage{times}  
\usepackage{helvet} 
\usepackage{courier}  
\usepackage[hyphens]{url}  
\usepackage{graphicx} 
\usepackage{natbib}  
\usepackage{caption} 
\usepackage{amsmath}
\usepackage{mathtools,amssymb}
\usepackage{bm}
\usepackage{lipsum,booktabs}
\usepackage{pgfplots}
\usepackage{multirow}
\pgfplotsset{compat=1.14}
\usepackage{wrapfig,lipsum,booktabs}
\title{Document-Level Relation Extraction with Reconstruction}
\author{
    Wang Xu \textsuperscript{\rm 1},
    Kehai Chen \textsuperscript{\rm 2},
    and Tiejun Zhao \textsuperscript{\rm {1}}
}
\affiliations{

    \textsuperscript{\rm 1} Harbin Institute of Technology, Harbin, China \\
    \textsuperscript{\rm 2} National Institute of Information and Communications Technology (NICT), Kyoto, Japan \\
    xuwang@hit-mtlab.net, khchen@nict.go.jp, tjzhao@hit.edu.cn
}

\begin{document}
\maketitle

\begin{abstract}
In document-level relation extraction (DocRE), graph structure is generally used to encode relation information in the input document to classify the relation category between each entity pair, and has greatly advanced the DocRE task over the past several years.
However, the learned graph representation universally models relation information between all entity pairs regardless of whether there are relationships between these entity pairs.
Thus, those entity pairs without relationships disperse the attention of the encoder-classifier DocRE for ones with relationships, which may further hind the improvement of DocRE.
To alleviate this issue, we propose a novel encoder-classifier-reconstructor model for DocRE.
The reconstructor manages to reconstruct the ground-truth path dependencies from the graph representation, to ensure that the proposed DocRE model pays more attention to encode entity pairs with relationships in the training.
Furthermore, the reconstructor is regarded as a relationship indicator to assist relation classification in the inference, which can further improve the performance of DocRE model.
Experimental results on a large-scale DocRE dataset show that the proposed model can significantly improve the accuracy of relation extraction on a strong heterogeneous graph-based baseline.
The code is publicly available at {https://github.com/xwjim/DocRE-Rec}.
\end{abstract}

\section{Introduction}
Graph structure plays an important role in the document relation extraction (DocRE) \cite{Christopoulou2019ConnectingTD,sahu-etal-2019-inter,Nan2020ReasoningWL,Tang2020HINHI}.
Typically, one unstructured input document is first organized as a structure input graph (i.e., homogeneous or heterogeneous graphs) based on syntactic trees, co-reference, or heuristics rules, thereby building relationships between entity pairs within and across multiple sentences of the input document.
Neural networks (i.e., graph network) are used to iteratively encode the structure input graph as a graph representation to model relation information in the input document.
The graph representation is fed into one classifier to classify the relation category between each entity pair, which has achieved the state-of-the-art performance in DocRE~\cite{Christopoulou2019ConnectingTD,Nan2020ReasoningWL}.

\begin{figure*}[!ht]
\centering
\includegraphics[scale=1]{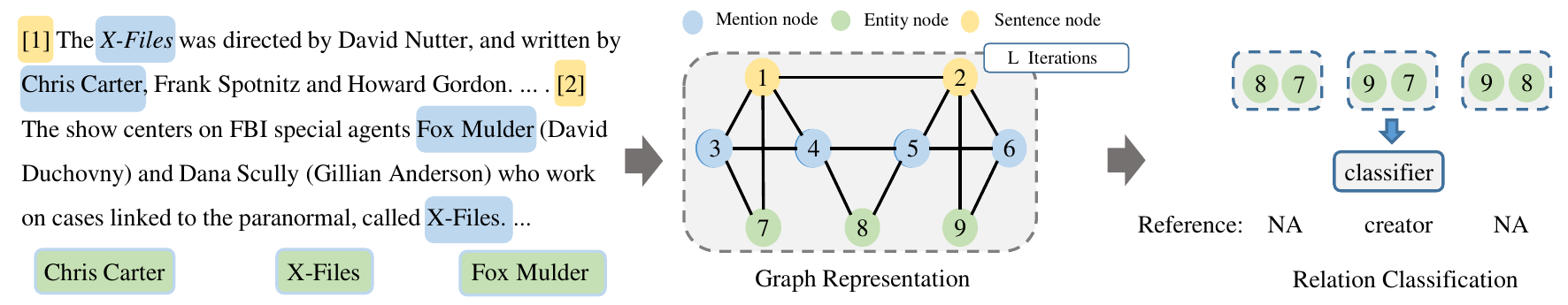} 
\caption{Heuristic rules are used to convert the input document into a heterogeneous graph. Then graph attention network is applied to learn the graph representation. Finally the node representations of entity pairs are used to classify their relationships.}
\label{fig1:background}
\end{figure*}

However, during the training of DocRE model, the graph representation universally encodes relation information between all entity pairs regardless of whether there are relationships between these entity pairs.
For example, Figure~\ref{fig1:background} shows three entities in an input document: \textit{X-Files}, \textit{Chris Carter}, and \textit{Fox Mulder}.
Intuitively, they are three entity pairs: \{\textit{X-Files}, \textit{Chris Carter}\}, \{\textit{X-Files}, \textit{Fox Mulder}\}, and \{\textit{Chris Carter}, \textit{Fox Mulder}\}.
The DocRE model learns the node representations of each entity pair to classify their relation.
As seen, there exists relationship between \{\textit{Chris Carter}, \textit{Fox Mulder}\} in the reference, indicating that there is naturally a reliable reasoning path from \textit{Chris Carter} to \textit{Fox Mulder}.
In comparison, there do not exist relationships between \{\textit{X-Files}, \textit{Chris Carter}\} and between \{\textit{X-Files}, \textit{Fox Mulder}\}, indicating that there are not reasoning paths between \{\textit{X-Files}-\textit{Chris Carter}\} or \{\textit{X-Files}, \textit{Fox Mulder}\}.
However, the learned graph representation models the three path dependencies universally and does not consider whether there is a path dependency between one target entity pair.
As a result, \{\textit{X-Files}, \textit{Chris Carter}\} and \{\textit{X-Files}, \textit{Fox Mulder}\} without relationships disperse the attention of the DocRE model for the learning of \{\textit{Fox Mulder}, \textit{Chris Carter}\} with relationship, which may further hinder the improvement of the DocRE model.

To alleviate this issue, we propose a novel reconstructor method to enable the DocRE model to model path dependency between one entity pair with the ground-truth relationship.
To this end, the reconstructor generates a sequence of node representations on the path from one entity node to another entity node and thereby maximizes the probability of its path if there is a ground-truth relationship between one entity pair and minimizes the probability otherwise.
This allows the proposed DocRE model to pay more attention to the learning of entity pairs with relationships in the training, thereby learning an effective graph representation for the subsequent relation classification.
Furthermore, the reconstructor is regarded as a relationship indicator to assist relation classification in the inference, which can further improve the performance of DocRE model.
Experimental results on a large-scale DocRE dataset show that the proposed method gained improvement of 1.7 F1 points over a strong heterogeneous graph-based DocRE model, especially outperformed the recent state-of-the-art LSR model for DocRE~\cite{Nan2020ReasoningWL}.

\section{Background}

In this section, based on \cite{Christopoulou2019ConnectingTD}'s work, we used heuristic rules to convert the input document into a heterogeneous graph without external syntactic knowledge.
Moreover, a graph attention network is used to encode the heterogeneous graph instead of the edge-oriented graph network~\cite{Christopoulou2019ConnectingTD}, thereby implementing a strong and general baseline for DocRE.

\subsection{Heterogeneous Graph Construction}
Formally, given an input document that consists of $L$ sentences \{$S^1$, $S^2$, $\cdots$, $S^L$\}, each of which is a sequence of words \{$x^{l}_{1}$, $x^{l}_{2}$, $\cdots$, $x^{l}_{J}$\} with the length $J$=$|S^l|$. 
A bidirectional long short-term memory (BiLSTM) reads word by word to generate a sequence of word vectors to represent each sentence in the input document.
Also, we apply the heterogeneous graph \cite{Christopoulou2019ConnectingTD} to the input document to build relationships between all entity pairs.
Specifically, the heterogeneous graph includes three defined distinct types of nodes: Mention Node, Entity Node, and Sentence Node.
For example, Figure \ref{fig1:background} shows an input document including two sentences (yellow color index) in which there are four mentions (blue color) and three entities (green color).
The representation of each node is the average of the words in the concept, thereby forming a set of node representations $\{\textbf{v}_{1}, \textbf{v}_{2}, \cdots, \textbf{v}_{N}\}$, where $N$ is the number of nodes.
For edge connections, there are five distinct types of edges between pairs of nodes following \cite{Christopoulou2019ConnectingTD}'s work, Mention-Mention(MM) edge, Mention-Sentence (MS) edge, Mention-Entity (ME) edge, Sentence-Sentence (SS) edge, Sentence-Sentence (SS) edge, Entity-Sentence (ES) edge respectively.
In addition, we add a Mention-Coreference (CO) edges between the two mentions which are referred to the same entity.
According to these above definitions, there is a $N\times N$ adjacency matrix $\mathbb{E}$ denoting edge connections.
Finally, the heterogeneous graph can be denoted as $G$=\{$\textbf{V}$, $\mathbb{E}$\}, to keep relation information between all entity pairs in the input document.

\subsection{Encoder}
\label{Sec3-1}
To learn an effective graph representation, we used the graph attention network~\cite{guo-etal-2019-attention} to encode the feature representation of each node in the heterogeneous graph.
Formally, given the outputs of all previous hop reasoning operations \{$\textbf{s}^{1}_{n}, \textbf{s}^{2}_{n}, \cdots, \textbf{s}^{l-1}_{n}$\},
they are concatenated and then transformed to a fixed dimensional vector as the input of the $l$ hop reasoning:
\begin{equation}
\begin{aligned}
\textbf{z}^l_n=\textbf{W}_e^l\cdot[\textbf{v}_n:\textbf{s}^{1}_{n}:\textbf{s}^{2}_{n}:\cdots:\textbf{s}^{l-1}_{n}],
\label{eq2:InputOfCNN}
\end{aligned}
\end{equation}
where $\textbf{s}^{l-1}_n$$\in$$\mathbb{R}^{d_0}$ and $\textbf{W}^l_e$$\in$$\mathbb{R}^{d_0\times (l\times d_0)}$.
Also, according to edge matrix $\mathbb{E}[n][a_c]$=k ($0 \le a_c < N, k>0)$, $C$ direct adjacent nodes of $\textbf{v}_n$ are $\{\textbf{z}^{l}_{a_1},\textbf{z}^{l}_{a_2}, \cdots, \textbf{z}^{l}_{a_C}\}$.
We then use the self-attention mechanism~\cite{NIPS2017_7181} to capture the feature information of $\textbf{v}_n$ between $\textbf{z}^l_n$ and $\{\textbf{z}^{l}_{a_1}, \textbf{z}^{l}_{a_1}, \cdots, \textbf{z}^{l}_{a_C}\}$:
\begin{equation}
\textbf{s}^{l}_{n}=\textup{softmax}(\frac{\textbf{z}^l_n\textbf{K}^{\top}}{\sqrt{d_{0}}})\textbf{V},
\label{eq5:Self-Attetion}
\end{equation}
where $\{\textbf{K}, \textbf{V}\}$ are key and value matrices that are transformed from the direct adjacent nodes representations $\{\textbf{z}^{l}_{a_1}, \textbf{z}^{l}_{a_1}, \cdots, \textbf{z}^{l}_{a_C}\}$ according to the edge type.

After performing $L$ hop reasonings, there is a sequence of annotations \{$\textbf{s}^{1}_{n}, \textbf{s}^{2}_{n}, \cdots, \textbf{s}^{L}_{n}$\} to encode relation information in the input document.
Finally, another no-linear layer is applied to integrate the reason information \{$\textbf{s}^{1}_{n}, \textbf{s}^{2}_{n}, \cdots, \textbf{s}^{L}_{n}$\} and the node information $\textbf{v}_n$:
\begin{equation}
\begin{aligned}
\textbf{q}_n=\textup{Relu}(\textbf{W}_{o}\cdot[\textbf{v}_n:\textbf{s}^1_n:\cdots:\textbf{s}^L_n]),
\label{eq3:ConvolutionOperation}
\end{aligned}
\end{equation}
where $\textbf{W}_{o}$$\in$$\mathbb{R}^{d_1 \times (d_0 \times (L+1))}$,
$\textbf{q}_n$$\in$$\mathbb{R}^{d_1}$.
As a result, the heterogeneous graph $G$ is represented as \{$\textbf{q}_1, \textbf{q}_2, \cdots, \textbf{q}_N$\}.

\subsection{Classifier}
Given the heterogeneous graph representation \{$\textbf{q}_1$, $\textbf{q}_2$, $\cdots$, $\textbf{q}_N$\}, two node representations of each entity pair are as the input to the classifier to classify their relationship. 
Specifically, the classifier is a multi-layer perceptron (MLP) layer with sigmoid function to calculate the relationship probability:
\begin{equation}
\begin{aligned}
R(r) = P(r|\{e_i, e_j\}) = \textup{sigmoid}(\textup{MLP}([\textbf{q}_i:\textbf{q}_j])).
\label{eq:combinatefeature}
\end{aligned}
\end{equation}

To train the DocRE model, the binary cross-entropy is used to optimize parameters of neural networks over the triple examples (subject, object, relation) on the training date set (including $T$ documents), that is, $\{\{e1^t_n, e2^t_n, r^t_{n}\}^{N_t}_{n=1}\}^T_{t=1}$:  
\begin{equation}
\begin{split}
Loss_c=-\frac{1}{\sum_{t=0}^TN_t}\sum_{t=1}^T\sum_{n=1}^{N_t}\{r^t_n\textup{log}(R(r^t_n))\\
+(1-r^t_n)\textup{log}(1-R(r^t_n))\},
\end{split}
\label{eq5:TrainingLoss}
\end{equation}
where $r^t_n \in\{0,1\}$ indicates whether the entity pair has relation label $r$ and $N_t$ is the number of relations in the $t$-th document. 


\section{Methodology}
\begin{figure}[h]
\centering
\includegraphics{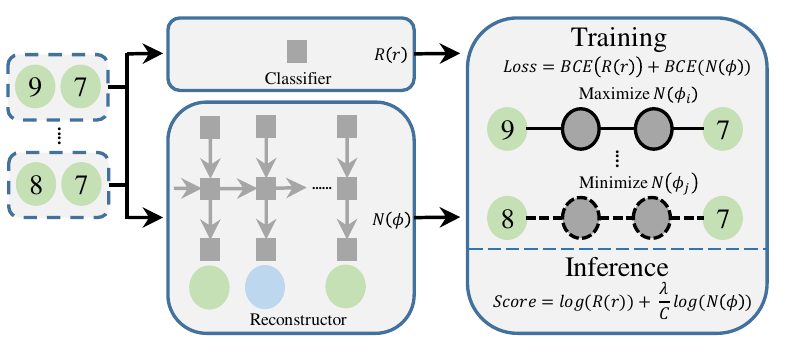} 
\caption{Model overview. The reconstructor manages to reconstruct the ground-truth path dependencies from the graph representation to ensure that the model to pay attention to model entity pairs with relationships. Furthermore, the reconstructor is regarded as a relationship indicator to assist relation classification in the inference.
}
\label{fig1:model}
\end{figure}

Intuitively, when a human understands a document with relationships, he or she often pays more attention to learn entity pairs with relationships rather than ones without relationships.
Motivated by this observation, we proposed a novel DocRE model with reconstruction (See Figure \ref{fig1:model}) to pay more attention to entity pairs with relationships, thus enhancing the accuracy of relationship classification.
\subsection{Meta Path of Entity Pair}
Generally, when there is a relationship between two entities, they should have one strong path dependency in the graph structure (or representation).
In comparison, when there is not a relationship between two entities, there is a weak path dependency.\footnote{If there is no path dependency between two target entities without a relationship, this may weaken the understanding of relationship information in the document.}
Thus, we explore to reconstruct the path dependency between each entity pair from the learned graph representation.
To this end, we first define three type paths between two entity nodes in the graph representation as reconstructed candidates according to the meta-path information~\cite{Sun2013MiningHI}.
\begin{enumerate}
\item[1)] Meta Path1 of Pattern Recognition: Two entities are connected through a sentence in this reasoning type. The relation schema is $EM \circ MM \circ EM$, for example node sequence \{7,3,4,8\} in Figure \ref{fig1:background}.

\item[2)] Meta Path2 of Logical Reasoning: the relation between two entities is indirectly established by a bridge entity. The bridge entity occurs in a sentence with the two entities separately. The relation schema is $EM \circ MM \circ CO \circ MM \circ EM$, for example node sequence \{7,3,4,5,6,9\} in Figure \ref{fig1:background}.

\item[3)] Meta Path3 of Coreference Reasoning: Coreference resolution must be performed first to identify target entities. A reference word refers to an entity that appear in the previous sentence. The two entities occur in the same sentence implicitly. The relation schema is $ES \circ SS \circ ES$, for example node sequence \{7,1,2,9\} in Figure \ref{fig1:background}.
\end{enumerate}
Actually, all the entity pairs have at least one of the three meta-paths. We select one meta-path type according to the priority, meta-path1 $>$ meta-path2 $>$ meta-path3.
Generally, several instance paths may exist corresponding to the meta path, we select the instance path that appears firstly in the document.

\subsection{Path Reconstruction}
For each entity pair, one instance path is selected as the supervision of the reconstruction of the path dependency.
In other words, there is only one supervision path $\phi_n$=\{$\textbf{v}_{b_1}$, $\textbf{v}_{b_2}$, $\cdots$, $\textbf{v}_{b_C}$\} between each target pair \{$e1_n$, $e2_n$\}, where $b_C$ is the number of nodes.

To reconstruct the path dependency of each entity pair, we model the reconstructor as the sequence generation.
Specifically, we use a LSTM to compute a path hidden state $\textbf{p}_{b_c}$ for each node $\textbf{q}_{b_{c-1}}$ on the path $\phi_n$:
\begin{equation}
\textbf{p}_{b_c}=\textup{LSTM}(\textbf{p}_{b_{c-1}}, \textbf{q}_{b_{c-1}}).
\label{eq9:lstminit}
\end{equation}
Note that $\textbf{p}_{b_0}$ is initialized as the transform of $\textbf{o}_{ij}$, since it plays a key role in classification.
$\textbf{p}_{b_c}$ is fed into a softmax layer to compute the probability of node $\textbf{v}_{b_c}$ on the path:
\begin{equation}
\begin{split}
\mathcal{P}(\textbf{v}_{b_c}|\textbf{v}_{<b_c})= \frac{\textup{exp}(\textbf{p}_{b_c} \textbf{W}_r \textbf{q}_{b_c}])}{\sum_{n}\textup{exp}(\textbf{p}_{b_c} \textbf{W}_r \textbf{q}_n])},
\end{split}
\label{eq7:nodeprobability}
\end{equation}
where $\textbf{W}_r \in \mathbb{R}^{d_1\times d_1}$. 
Also, there is a set of node probabilities  \{$\mathcal{P}(\textbf{v}_{b_1}|\textbf{v}_{<b_1})$, $\mathcal{P}(\textbf{v}_{b_2}|\textbf{v}_{<b_2})$, $\cdots$, $\mathcal{P}(\textbf{v}_{b_C}|\textbf{v}_{<b_C})$\} for the path $\phi_n$.
Finally, the probability of this path $\phi_n$ is computed:
\begin{equation}
\mathcal{N}(\phi_n)= \prod^{C}_{c=1}(\mathcal{P}(\textbf{v}_{b_c}|\textbf{v}_{<b_c}).
\label{eq8:pathprobability}
\end{equation}

\subsection{Training with Reconstruction Loss}


We use the reconstructed path probability to compute an additional reconstruction loss over the triple examples of the training data set $\{\{e1^t_n, e2^t_n, r^t_{n}\}^{N_t}_{n=1}\}^T_{t=1}$:
\begin{equation}
\begin{split}
Loss_r=-\frac{1}{\sum_{t=0}^TN_t}\sum_{t=1}^T\sum_{n=1}^{N_t}\{r^t_n\textup{log}\mathcal{N}(\phi_n)\\
+(1-r^t_n)\textup{log}(1-\mathcal{N}(\phi_n)\},
\label{eq9:loss2}
\end{split}
\end{equation}
where $r^t_n$ is one of \{0,1\}, that is, we maximize the probability of the path $\mathcal{N}(\phi_n)$ if the entity pair has relation, and minimize the probability otherwise.
To simplify the Eq.\eqref{eq9:loss2}, we use $\prod_{c=1}^{C}(1-P_{b_c})$ to replace with the $(1-\mathcal{N}(\phi_n))$, where $P_{b_c}$=$\mathcal{P}(\textbf{v}_{b_c}|\textbf{v}_{<b_c})$. 
The reconstruction loss is modified as Eq.\eqref{eq10:loss2new}:
\begin{equation}
\begin{split}
Loss_r=-\frac{1}{\sum_{t=0}^TN_t}\sum_{t=1}^T\sum_{n=1}^{N_t}\{\sum_{b_c=1}^{b_C}\{(r^t_n\textup{log}P_{b_c})\\
+(1-r^t_n)\textup{log}(1-P_{b_c})\}\}.
\label{eq10:loss2new}
\end{split}
\end{equation}
Finally, the reconstructor loss and the existing classification loss in Eq.\eqref{eq5:TrainingLoss} is added as the training objective of the proposed DocRE model:
\begin{equation}
\begin{aligned}
Loss=Loss_c+Loss_r.
\label{eq11:LossFunction}
\end{aligned}
\end{equation}%

\subsection{Inference with Path Reconstruction}
Intuitively, the proposed reconstructor encourages the DocRE model to pay more attention to model entity pairs with ground-truth relationships.
Furthermore, we maximized the path probability between one entity pair if there is indeed a relation and we minimized it otherwise when computing the reconstruction loss in Eq.\eqref{eq10:loss2new}.
In other words, the higher the probability of this path is, the greater the likelihood of a relationship between the entity pair is.
Naturally, we treat this path probability as a relational indicator to assist relation classification in the inference:
\begin{equation}
\begin{split}
S(r) = \textup{log}(R(r)) + \lambda \cdot \frac{1}{C}\sum_{b_c=1}^{b_C}\textup{log}(P_{b_c}),
\label{eq13:relationscores}
\end{split}
\end{equation}%
where $\lambda$ is a hyper-parameter to control the importance of reconstruction probability in the inference. 
\begin{table*}[h]
\begin{center}
\scalebox{.92}{
\begin{tabular}{l|l|l|l|l|l}
\multirow{2}{*}{Groups} & \multicolumn{1}{c|}{\multirow{2}{*}{Methods}} & \multicolumn{2}{c|}{Dev} & \multicolumn{2}{c}{Test} \\ \cline{3-6} 
                       & \multicolumn{1}{c|}{}                       & Ign F1        & F1       & Ign F1        & F1        \\ \hline
\multirow{10}{*}{w/o BERT}    
& CNN$^*$~\cite{yao-etal-2019-docred}           & 41.58 & 43.45 & 40.33 & 42.26 \\ 
& BiLSTM$^*$~\cite{yao-etal-2019-docred}        & 48.87 & 50.94  & 48.78 & 51.06          \\ 
& ContexAware$^*$~\cite{yao-etal-2019-docred}   & 48.94 & 51.09 & 48.40 & 50.07 \\
                       & GCNN$^\dag$~\cite{sahu-etal-2019-inter}          & 46.22 & 51.52 & 49.59 & 51.62         \\ 
                       &  EoG$^\dag$~\cite{Christopoulou2019ConnectingTD}  & 45.94 & 52.15 & 49.48 & 51.82        \\ 
                       & GAT$^\dag$~\cite{Velickovic2018GraphAN}          & 45.17 & 51.44 & 47.36 & 49.51          \\ 
                       &  AGGCN$^\dag$~\cite{guo-etal-2019-attention}      & 46.29 & 52.47 & 48.89 & 51.45        \\ 
                       &  LSR$^*$~\cite{Nan2020ReasoningWL}             & 48.82 & 55.17 & 52.15 & 54.18        \\ \cline{2-6} 
                       & HeterGSAN                                 & 52.17 & 54.40 & 52.07 & 53.52     \\ 
                       & \;\;\;+Reconstruction                  & \textbf{54.27} & \textbf{56.22} & \textbf{53.27} & \textbf{55.23}   \\ \hline \hline
\multirow{7}{*}{w/ BERT}     
                       &  BERT$^*$~\cite{Wang2019FinetuneBF}            & -     & 54.16 & -     & 53.20      \\ 
                       &  Two-Phase BERT$^*$~\cite{Wang2019FinetuneBF}  & -     & 54.42 & -     & 53.92          \\ 
                       & BERT+LSR$^*$~\cite{Nan2020ReasoningWL}        & 52.43 & 59.00 & 56.97 & 59.05        \\ \cline{2-6} 
                       &    HeterGSAN                                 & 52.17 & 54.40 & 52.07 & 53.52       \\ 
                       &  \;\;\;+BERT                           & 57.00 & 59.13 & 56.21 & 58.54         \\ 
                       &  \;\;\;\;\;\;+Reconstruction             & \textbf{58.13} & \textbf{60.18} & \textbf{57.12} & \textbf{59.45}   \\ 
\end{tabular}}
\end{center}
\caption{Results on the development set and the test set. Results with $*$ are reported in their original papers. Results with $\dag$ are reported in \cite{Nan2020ReasoningWL}. Bold results indicate the best performance of the current method.}
\label{tab1:result} 
\end{table*}

\section{Experiments}
\subsection{Setup}
The proposed methods were evaluated on a large-scale human-annotated dataset for document-level relation extraction~\cite{yao-etal-2019-docred}. 
DocRED contains 3,053 documents for the training set, 1,000 documents for the development set, and 1,000 documents for the test set, totally with 132,375 entities, 56,354 relational facts, and 96 relation types. 
More than 40\% of the relational facts require the reading and reasoning over multiple sentences.
Following settings of~\cite{Nan2020ReasoningWL}'s work, we used the GloVe embedding (100d) and BiLSTM (128d) as word embedding and encoder.
The hop number $L$ of the encoder was set to 2. 
The learning rate was set to 1e-4 and we trained the model using Adam as the optimizer.
For the BERT representations, we used uncased BERT-Based model (768d) as the encoder and the learning rate was set to $1e^{-5}$
For evaluation, we used $F_1$ and Ign $F_1$ as the evaluation metrics. 
Ign $F_1$ denotes $F_1$ score excluding relational facts shared by the training and development/test sets.
In particular, the predicted results were ranked by their confidence and traverse this list from top to bottom by $F1$ score on development set, and the score value corresponding to the maximum $F1$ is picked as threshold $\theta$.
All hyper-parameters were tuned based on the development set.
In addition, the results on the test set were evaluated through CodaLab\footnote{\url{https://competitions.codalab.org/competitions/20717}}.
%

\subsection{Baseline Systems}
According to Section 2, there is a baseline heterogeneous-based graph self-attention network model (HeterGSAN).
Also, there are some recent DocRE methods as our comparison systems:

$\bullet$  \textbf{Sequence-based Models:} 
These models used different neural architectures to encode sentences in the document, including including convolution neural networks (CNN) \cite{yao-etal-2019-docred}, bidirectional LSTM (BiLSTM) \cite{yao-etal-2019-docred} and Context-Aware LSTM \cite{yao-etal-2019-docred}.

$\bullet$  \textbf{Graph-based Models.}
GCNN~\cite{sahu-etal-2019-inter}, GAT~\cite{Velickovic2018GraphAN}, AGGCN~\cite{guo-etal-2019-attention} constructed the graph from syntactic parsing and sequential information, or non-local dependencies from coreference resolution and other semantic dependencies, and then uses the GCN based method to calculate the node embedding. 
EoG~\cite{Christopoulou2019ConnectingTD} defined several node types and edges to construct a heterogeneous graph of the input document without external syntactic knowledge. EoG uses an iterative algorithm to learn new edge representations between different nodes in the heterogeneous graph and classify relationships between entity pairs. 
Instead of constructing a static graph representation, LSR~\cite{Nan2020ReasoningWL} empowered the relational reasoning across sentences by automatically inducing the latent document-level graph.


$\bullet$  \textbf{BERT.} 
It applied a pre-trained language model to learn the representations of the input document~\cite{Wang2019FinetuneBF,Devlin2019BERTPO}.
Furthermore, it used a two-phase training process to enhance the performance of DocRE model. 
Specifically, it first predicts whether a pair of entities has a relation or not and classifies the relation for each entity pair.

\subsection{Main Results}
Table \ref{tab1:result} presents the detailed results on the development set and the test set of DocRED. 
As seen, our baseline HeterGSAN model achieved 53.52 $F1$ score on the test set and outperformed the EoG model which is also a heterogeneous-based graph DocRE model by 1.7 points in terms of $F_1$.
Meanwhile, HeterGSAN is consistently superior to the most of comparison methods, including CNN, BiLSTM, ContextAware, GCNN, GAT, and AGGCN.
This indicates that the graph self-attention network can give a strong baseline in the heterogeneous-based methods of DocRE.
HeterGSAN+reconstruction achieved 55.23 $F1$, which outperformed the baseline HeterGSAN by 1.71 $F_1$ score.
In particular, HeterGSAN+reconstruction outperformed the existing state-of-the-art LSR model by 1.05 $F_1$ score, which is a new state-of-the-art result on the DocRED dataset without the pre-trained model (BERT).
This means that the proposed reconstructor is beneficial to encode relation information in the input document, thereby enhancing the relation extraction.


In addition, we evaluated the proposed HeterGSAN model with a pre-trained language model as shown in Table~\ref{tab1:result}.
First, HeterGSAN+BERT model consistently outperformed the comparison BERT model, Two-Phase BERT model, BERT+LSR model.
This confirms the effectiveness of the BERT method, which we believe makes the evaluation convincing.
Moreover, HeterGSAN+BERT+Reconstruction model outperformed HeterGSAN+BERT model by 0.91 $F_1$ score, indicating that our approach is complementary to BERT, and combining them is able to further improve the accuracy of relation extraction.
Meanwhile, HeterGSAN+BERT+Reconstruction model ($F_1$ 59.45) outperformed BERT+LSR model ($F_1$ 59.05) by 0.40 $F_1$ score on the test set, which is a new state-of-the-art result.

\subsection{Effect of Reconstruction}
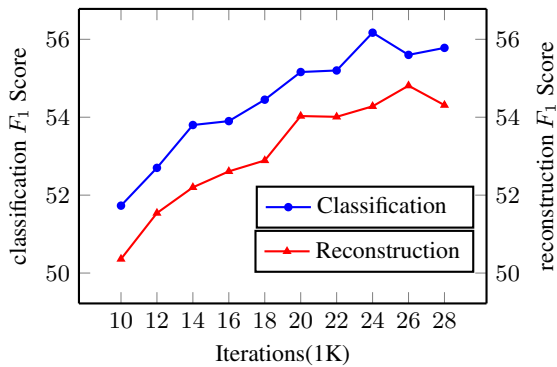
\begin{figure}[h]
	\begin{center}
		\pgfplotsset{height=5.5cm,width=8.5cm,compat=1.14,every axis/.append style={thick},every axis legend/.append style={at={(1,0.13)}},legend columns=1}
		\begin{tikzpicture}
		\tikzset{every node}=[font=\small]
		\begin{axis}
		[width=7cm,enlargelimits=0.13, tick align=inside, 
		xtick={10,12,14,16,18,20,22,24,26,28},
		ymin=50,
		ymax=56,
		axis y line*=left,
		legend style={at={(0.97,0.4)}},
		ylabel={classification $F_1$ Score},xlabel={Iterations(1K)},font=\small]
	
		\addplot [sharp plot,mark=*,mark size=1.2pt,mark options={solid,mark color=blue}, color=blue] coordinates
		{ (10,51.73)(12,52.70)(14,53.80)(16,53.90)(18,54.45)(20,55.16)(22,55.20)(24,56.17)(26,55.6)(28,55.78)};
		\addlegendentry{\small Classification\;\;}
		\end{axis}
		
		\begin{axis}
		[width=7cm,enlargelimits=0.13, tick align=inside, 
		xtick={10,12,14,16,18,20,22,24,26,28},
		ymin=50,
		ymax=56,
		axis y line*=right,
		axis x line=none,
		ylabel style = {align=center},
		ylabel={reconstruction $F_1$ Score},
		legend style={at={(0.97,0.25)}},
		font=\small]
		\addplot [sharp plot,mark=triangle*,mark size=1.2pt,mark options={solid,mark color=red}, color=red] coordinates
		{ (10,50.36)(12,51.54)(14,52.20)(16,52.61)(18,52.89)(20,54.03)(22,54.01)(24,54.28)(26,54.81)(28,54.31)};
		\addlegendentry{\small Reconstruction}
		\end{axis}
		
		\end{tikzpicture}
		\caption{\label{fig:learningcurves}Learning curves of classification (left y-axis) and reconstruction (right y-axis) performances (in F1 scores) on the development set during the training.}
	\end{center}
\end{figure}
To valid the effect of reconstruction, Figure \ref{fig:learningcurves} showed learning curves of classification and reconstruction performances (in F1 scores) on the development set during the training. 
For reconstruction, we used the reconstructor to generate the source path for each entity pair and calculated the probability of the reconstructed path to indicate how much there is a relationship. 
As seen, the reconstruction $F1$ scores went up with the improvement of reconstruction over time. 
When the classification performance reached a peak at iteration 24K, the proposed model achieved a balance between classification and reconstruction scores. Therefore, we use the trained model at iteration 24K in Table~\ref{tab1:result}.
\subsection{Ablation in Training and Inference}
\begin{table}[h]
	\begin{center}
	\scalebox{.92}{
		\begin{tabular}{c|cc|c|c}
			 & \multicolumn{2}{c|}{Reconstructor used in} & \multicolumn{2}{c}{Metric} \\ \hline
			& Training        & Inference        & Ign F1        & F1     \\ \hline
			\#1 & $\times$        & $\times$       &  52.17    & 54.40   \\
			\#2 & \checkmark      & $\times$       &  53.69    & 55.66    \\
			\#3 & \checkmark      & \checkmark     &  54.27    & 56.22    \\ 
		\end{tabular}}
	\end{center}
	\caption{\label{tab3:contributionanalysis}Ablation of Reconstructor in training and inference.}
\end{table}
To further explore the effect of Reconstructor, we incrementally introduced it into the training and inference phases in turn.
Table~\ref{tab3:contributionanalysis} shows the results of the ablation experiment on the development set.  
As seen, when Reconstructor was only introduced into the training phase (\#2), there was 1.26 $F_1$ improvement over the baseline HeteGASN model (\#1) in which there are not Reconstructor in the training and inference phases.
Moreover, Reconstructor was introduced into the inference as a relation indicator to assist relation classification, that is, there are Reconstructor in both training and inference contain the Reconstructor (\#3), 
As a result, there gained 0.56 $F_1$ further improvement. 
This shows that the proposed Reconstructor can not only encode relation information of the input document efficient but also indicate how much there is a relationship, to enhance relation classification between entity pair.

\subsection{Ablation of Reconstruction Loss}
In the reconstruction phase, we maximized (max) the path probability if the entity pair has the ground-truth relationship and minimized (min) the path probability otherwise.
Therefore, we performed the ablation of the above two reconstruction paths.
Specifically, we gradually introduced them into the proposed HeterGSAN with Reconstruction to verify the effect of two reconstruction paths, as shown in Table~\ref{tab4:pathablation}.
Here, ``relation" denotes entity pairs with ground-truth relationships while ``no-relation" denotes entity pairs without ground-truth relationships.
As seen, when one of ``no-relation" (\#2) and ``relation" (\#3) entity pairs were used to compute the reconstruction loss, their $F_1$ scores were better than the baseline HeterGSAN (\#1).
This means that reconstructing one of two paths is beneficial to improve the performance of DocRE model.
Meanwhile, ``relation" (\#3) was superior to ``no-relation" (\#2).
In particular, both of them can complement each other to further improve $F_1$ score (\#4).
This indicates that two path reconstruction methods help the DocRE model capture more diverse useful information from the input document.
\begin{table}[h]
	\begin{center}
		\begin{tabular}{c|cc|c|c}
			& \multicolumn{2}{c}{entity pair} & \multicolumn{2}{c}{Metric} \\ \cline{2-5}
			& relation        & no-relation          & Ign F1        & F1     \\ \hline
			\#1 & --              & --      &  52.17    & 54.40    \\
			\#2 & --              & min     &  53.29    & 55.04    \\
			\#3 & max             & --      &  53.53    & 55.55    \\
			\#4 & max             & min     &  54.27    & 56.22    \\ 
		\end{tabular}
	\end{center}
	\caption{\label{tab4:pathablation}Ablation experiments of reconstruction loss for the proposed HeterGSAN+Reconstruction model.}
\end{table}
\subsection{Effect of Path Probability in Inference}
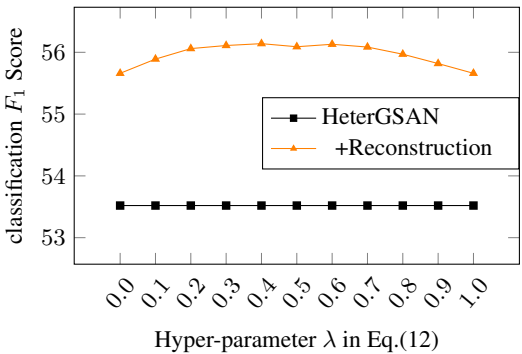
\begin{figure}[h]
	\begin{center}
		\pgfplotsset{height=5.0cm,width=8.5cm,compat=1.14,every axis/.append style={thick},every axis legend/.append style={at={(0.8,1.13)}},legend columns=1}
		\begin{tikzpicture}
		\tikzset{every node}=[font=\small]
		\pgfplotsset{set layers}
		\begin{axis}
		[width=7.5cm,enlargelimits=0.13, tick align=inside, xticklabels={ $0.0$, $0.1$,$0.2$, $0.3$, $0.4$,$0.5$,$0.6$,$0.7$,$0.8$,$0.9$,$1.0$},
		xtick={0.0,0.1,0.2,0.3,0.4,0.5,0.6,0.7,0.8,0.9,1.0},
		ymin=53,
		ymax=56.3,
		x tick label style={rotate=50},
		legend style={at={(1.0,0.65)}},
		ylabel={classification $F_1$ Score},xlabel={Hyper-parameter $\lambda$ in Eq.\eqref{eq13:relationscores}},font=\small]

		\addplot+ [sharp plot,mark=square*,mark size=1.2pt,mark options={solid,mark color=black}, color=black] coordinates
		{ (0.0,53.52)(0.1,53.52)(0.2,53.52)(0.3,53.52)(0.4,53.52)(0.5,53.52)(0.6,53.52)(0.7,53.52)(0.8,53.52)(0.9,53.52)(1.0,53.52)};
		\addlegendentry{\small HeterGSAN\;\;\;\;\;\;\;\;\;\;}
		
		\addplot+ [sharp plot,mark=triangle*,mark size=1.2pt,mark options={solid,mark color=orange}, color=orange] coordinates
		{ (0.0,55.66)(0.1,55.89)(0.2,56.06)(0.3,56.11)(0.4,56.14)(0.5,56.09)(0.6,56.13)(0.7,56.085)(0.8,55.969)(0.9,55.818)(1.0,55.66)};
		\addlegendentry{\small +Reconstruction}
		

		\end{axis}
		
		\end{tikzpicture}
		\caption{\label{fig:hyperlambda} Classification $F_1$ scores of different hyper-parameter $\lambda$ for the reconstructed path probability of DocRE models (HeterGSAN and +Reconstruction) in inference.}
	\end{center}
\end{figure}
\begin{figure*}[t]
\centering
\includegraphics{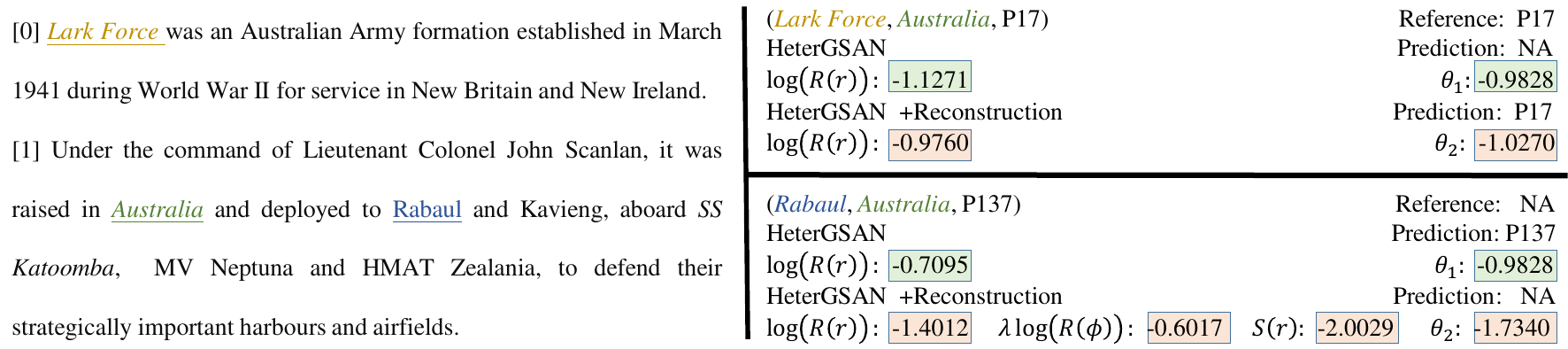} 
\caption{Case Study}
\label{fig:casestudy}
\end{figure*}
In inference, the reconstructor is regarded as a relationship indicator to assist relation classification. 
The hyper-parameter $\lambda$ in Eq.\eqref{eq13:relationscores} keeps a trade-off between the classification scores and the construction scores when classifying the relation of each entity pair.
Figure \ref{fig:hyperlambda} shows classification $F_1$ scores of different hyper-parameter $\lambda$ for the reconstructed path probability of HeterGSAN and +Reconstruction models in inference.
As seen, $F_1$ scores of +Reconstruction model increased with the increasing of $\lambda$ until 0.4, indicating that the probability of reconstructed path is useful for improving the relation classification.
Subsequently, larger values of $\lambda$ reduced the $F_1$ scores, suggesting that excessive biased path information may be weak at keeping the gained improvement.
Therefore, we set the hyper-parameter $\lambda$ to 0.4 to control the effect of reconstructed path information in our experiments (Table~\ref{tab1:result}).

\subsection{Evaluating Different Meta Path}
To evaluate our defined three candidate meta paths, we divide entity pairs of the same type of meta path in the development set to three groups, for example, ``MP1" indicates that the path representation of entity pairs are from the defined Meta Path1 (See section 3.1) during the reconstruction.   
Table~\ref{tab5:pathlen} showed $F_1$ scores of three groups (MP1, MP2, and MP3) for HeterGSAN and +Reconstruction models.
As seen, $F_1$ scores of +Reconstruction outperformed that of HeterGSAN in all three groups.
This means that our defined meta paths can efficient capture path dependency between entity pairs in the reconstruction processing.
\begin{table}[h]
    \begin{center}
        \begin{tabular}{l|c|c|c}
                        & MP1 & MP2 & MP3 \\ \hline
        HeterGSAN~(\%)       & 60.67                      & 50.29 & 46.30 \\
        \;\;\;\;+Reconstruction~(\%) & 61.73                      & 52.19 & 47.57 \\ 
        \end{tabular}
    \end{center}
    \caption{\label{tab5:pathlen}$F_1$ scores of three groups (MP1, MP2, and MP3) with different meta paths.}
\end{table}
\subsection{Path Attention Scores}
To study how the reconstructor (Rec) affect the distribution of attention scores along the path in the HeterGSAN, we divided attention scores into five intervals (i.e, 0-0.2, 0.2-0.4, etc) and showed the percent of attention distribution on HeterGSAN and +Reconstruction on the development set as shown in the Table \ref{tab9:pathscores}.
The attention scores of HeterGSAN are mainly concentrated in interval 0-.2, which may indicate the hypothesis of universally learning relationship information. 
Thus, +Reconstruction significantly reduced the percent of attention scores in interval 0-.2 and increased the percent of remaining intervals with higher attention scores. 
This means that the reconstructor guides the DocRE model to pay more attention to model meta-path dependencies for the ground-truth relationships.
\begin{table}[h]
    \begin{center}
        \begin{tabular}{l|c|c|c|c|c}
                       & 0-.2& .2-.4&.4-.6& .6-.8&.8-1.0 \\ \hline
        HeterGSAN(\%)       & 84.43&5.28&2.28& 2.53& 5.48    \\
        \;\;\;\;+Rec~(\%)   & 69.04&9.21&10.78& 1.34& 9.63    \\ 
        \end{tabular}
    \end{center}
    \caption{\label{tab9:pathscores}Changes of the distribution of path attention scores}
\end{table}

\subsection{Ablation of Different Meta-Paths}
we reconstruct one of three meta-paths (MP1, MP2 and MP3) in each DocRE model and not consider the reconstructor in inference. 
The results are as follows in Table \ref{tab10:ablation_of_metapath}.
First, reconstruction of each meta-path is beneficial to enhance the DocRE model, confirming our motivation. 
Thus, the improved range of each meta-path is in descending order: MP1, MP2, MP3, confirming the priority for the reconstruction meta-path in Sec 3.1. 
It is a statistic that the percentage of MP1, MP2, and MP3 are 22.39\%, 23.15\%, and 54.46\%. 
Then, when two different meta-paths are considered, their F1 values are higher than the single path which is reconstructed, indicating that more ground-truth path relationships are reconstructed to enhance the training of the DocRE model. 
Similarly, considering three meta-paths gain the highest F1 on development/test sets.
\begin{table}[h]
    \begin{center}
    \scalebox{.95}{
        \begin{tabular}{c|c|c||c|c|c}
        \begin{tabular}[c]{@{}c@{}}type of\\ meta-path\end{tabular} & \begin{tabular}[c]{@{}c@{}}Dev\\ F1\end{tabular} & \begin{tabular}[c]{@{}c@{}}Test\\ F1\end{tabular} & \begin{tabular}[c]{@{}c@{}}type of\\ meta-path\end{tabular} & \begin{tabular}[c]{@{}c@{}}Dev\\ F1\end{tabular} & \begin{tabular}[c]{@{}c@{}}Test\\ F1\end{tabular} \\ \hline
        None                                                        & 54.40                                            & 53.52                                             & MP1\&MP2                                                       & 55.26                                            & 54.40                                             \\
        MP1                                                          & 54.79                                            & 54.22                                             & MP1\&MP3                                                       & 55.12                                            & 54.37                                             \\
        MP2                                                          & 54.78                                            & 54.20                                             & MP2\&MP3                                                       & 54.96                                            & 54.28                                             \\
        MP3                                                          & 54.54                                            & 53.88                                             & All                                                         & 55.66                                            & 54.91                                             \\ 
        \end{tabular}}
    \end{center}
    \caption{\label{tab10:ablation_of_metapath}Ablation experiments of different Meta-Paths.}
\end{table}

\subsection{Case Study}
Figure \ref{fig:casestudy} shows a case study of HeterGSAN and +Reconstruction models. 
For the entity pair \{\textit{Lark Force}, \textit{Australia}\}, HeterGSAN classified its relation to ``NA" which is inconsistent with the Reference ``P17" because of its classifier score -1.1271 is less than the threshold $\theta_1$ -0.9828.
In comparison, the classifier score of +Reconstruction classified its relation to ``P17" which is consistent with the Reference ``P17" because of its classifier score -0.9760 was greater than the threshold $\theta_2$ -1.0270.
This means that the proposed Reconstructor can better guild the training of DocRE model.
For another entity pair \{\textit{Rabaul}, \textit{Australia}\}, the classifier scores of HeterGSAN and +Reconstruction models were greater than $\theta_1$ and $\theta_2$, respectively.
However, they gained a relation category P137 which is inconsistent with the Reference ``NA".
When the path score -0.6017 was considered in the inference, +Reconstruction classified its relation to ``NA" which is consistent with the Reference ``NA".
This indicates that the inference with Reconstructor can further improve the accuracy of relation classification.

\section{Related Work}
\textbf{DocRE} 
Early efforts focus on classifying relationships between entity pair within a single sentence or extract entity and relations jointly in a sentence \cite{zeng-etal-2014-relation,wang-etal-2016-relation,wei2019novel,Song_2019}. 
These approaches do not consider interactions across mentions and ignore relations expressed across sentence boundaries. 
Recently, the extraction scope has been expanded to the entire document in the biomedical domain by only considering a few relations among chemicals \cite{DBLP:journals/tacl/PengPQTY17,quirk-poon-2017-distant,DBLP:journals/corr/abs-1810-05102,zhang-etal-2018-graph,Christopoulou2019ConnectingTD}. 
In particular, \citet{yao-etal-2019-docred} proposed a large-scale human-annotated DocRED dataset.
The dataset requires understanding a document and performing multi-hop reasoning and several works \cite{Wang2019FinetuneBF,Nan2020ReasoningWL} have been done on the dataset.

\textbf{Reconstruction}
Reconstructor was used to solve the problem that translations generated by neural network translation (NMT) often lack adequacy \cite{nmt-rec,nmt-chen}. \cite{nmt-chen} reconstructs the monolingual corpora with two separate source-to-target and target-to-source NMT models. \cite{nmt-rec} aims at enhancing adequacy of unidirectional (i.e., source-to-target) NMT via a target-to-source objective on parallel corpora.
Besides, \cite{gpt-gnn} uses reconstructor to pre-train a graph neural network on the unlabeled data with self-supervision to reduce the cost of labeled data.

\section{Conclusion}
This paper proposed a novel reconstruction method to guide the DocRE model to pay more attention to the learning of entity pairs with the ground-truth relationships, thereby learning an effective graph representation to classify relation category.
In inference, the reconstructor is further regarded as a relation indicator to assist relation classification between entity pair. 
Experimental results on a large-scale DocRED dataset show that our method can greatly advance the DocRE task.
In the future, we will explore more information related to relationship classification in the input document, for example, syntax constraint~\cite{AAAI1816060}, diverse information~\cite{9097389}, and knowledge reasoning~\cite{Cohen2020Scalable}.  

\section*{Acknowledgments}
We are grateful to the anonymous reviewers, senior program Committee and area chair for their insightful comments and suggestions.
The corresponding authors are Kehai Chen and Tiejun Zhao.
This work is supported by the National Key R\&D Program of China (No. 2018YFC0830700) and Huawei Technologies CO., Ltd (No. YBN2019115122).

\bibliography{main}

\end{document}